\documentclass[pdflatex,sn-mathphys-num]{sn-jnl}

\usepackage{graphicx}%
\usepackage{multirow}%
\usepackage{amsmath,amssymb,amsfonts}%
\usepackage{amsthm}%
\usepackage{mathrsfs}%
\usepackage[title]{appendix}%
\usepackage{xcolor}%
\usepackage{textcomp}%
\usepackage{manyfoot}%
\usepackage{rotating}%
\usepackage{booktabs}%
\usepackage{algorithm}%
\usepackage{algorithmicx}%
\usepackage{algpseudocode}%
\usepackage{listings}%
\usepackage{array}%

\usepackage{mathtools} 
\usepackage{comment}

\theoremstyle{thmstyleone}%
\newtheorem{theorem}{Theorem}
\newtheorem{lemma}[theorem]{Lemma}
\newtheorem{corollary}[theorem]{Corollary}   

\theoremstyle{thmstyletwo}%

\theoremstyle{thmstylethree}%
\newtheorem{definition}{Definition}%

\newcommand{\ind}{\perp\!\!\!\!\perp}

\newcommand{\repname}{}  
\newtheorem*{reptheoreminner}{\repname}

\newenvironment{replemma}[1]{%
  \renewcommand{\repname}{Lemma~\ref{#1}}%
  \begin{reptheoreminner}%
}{%
  \end{reptheoreminner}%
}

\begin{document}

\title[Article Title]{Performance Estimation in Binary Classification Using Calibrated Confidence}


\author*[1]{\fnm{Juhani} \sur{Kivimäki}}\email{juhani.kivimaki@helsinki.fi}

\author[2]{\fnm{Jakub} \sur{Białek}}\email{jakub.bialek@nannyml.com}

\author[2]{\fnm{Wojtek} \sur{Kuberski}}\email{wojtek.kuberski@nannyml.com}

\author[1]{\fnm{Jukka} \spfx{K.} \sur{Nurminen}}\email{jukka.k.nurminen@helsinki.fi}

\affil*[1]{\orgname{University of Helsinki}, \orgaddress{\street{PL 64}, \postcode{00014}, \city{Helsinki},  \country{Finland}}}

\affil[2]{\orgname{NannyML}, \orgaddress{\street{Diestsesteenweg 50/20}, \city{Leuven}, \country{Belgium}}}     


\abstract{Model monitoring is a critical component of the machine learning lifecycle, safeguarding against undetected drops in the model's performance after deployment. Traditionally, performance monitoring has required access to ground truth labels, which are not always readily available. This can result in unacceptable latency or render performance monitoring altogether impossible. Recently, methods designed to estimate the accuracy of classifier models without access to labels have shown promising results. However, there are various other metrics that might be more suitable for assessing model performance in many cases. Until now, none of these important metrics has received similar interest from the scientific community. In this work, we address this gap by presenting \textit{Confidence-based Performance estimation} (CBPE), a novel method that can estimate any binary classification metric defined using the confusion matrix. In particular, we choose four metrics from this large family: accuracy, precision, recall, and F$_1$, to demonstrate our method. CBPE treats the elements of the confusion matrix as random variables and leverages calibrated confidence scores of the model to estimate their distributions. The desired metric is then also treated as a random variable, whose full probability distribution can be derived from the estimated confusion matrix. CBPE is shown to produce estimates that come with strong theoretical guarantees and valid confidence intervals. }


\maketitle

\section{Introduction}\label{sec:intro}

In recent years, the problem of \textit{unsupervised accuracy estimation}~\citep{kivimaki:2025} has drawn increasing attention. The task is to estimate the accuracy of a given machine learning (ML) model under some potentially shifted data distribution, when measuring the accuracy directly is either inconvenient or altogether impossible due to a lack of access to ground truth (GT) labels. Such situations are surprisingly common in real-life use cases where querying the labels for in-production data might be prohibitively expensive, or the labels might become available only after unacceptable latency~\citep{krawczyk:2017}. However, most users would still like to remain informed about the predictive performance of the deployed model.

A notable shortcoming of most earlier approaches has been that they all focus solely on predicting the accuracy of a given model. However, accuracy can be an uninformative or misleading metric in many cases, such as when there is a severe class imbalance or when different types of errors incur different costs~\citep{bekkar:2013}. In this work, we offer a remedy by expanding the scope of unsupervised accuracy estimation into \textit{unsupervised performance estimation} and introducing \textit{Confidence-based performance estimation} (CBPE). Our approach applies to any classification metric, which can be defined using the elements of the \textit{Confusion Matrix}. Since there is a plethora of such metrics, we introduce our method using only four common members of this family (accuracy, precision, recall, and F$_1$) as examples. 

Another novelty in our approach is that we explore our estimators in a monitoring setting, whereas earlier approaches have been examined on the dataset level. We argue that the monitoring setting is more important for practical use cases where one typically has access only to a limited amount of instances from the target distribution instead of a full dataset with potentially tens of thousands of samples. Understandably, estimating model performance is a harder task in the monitoring setting and estimators shown to perform well on the dataset level might not work as well in a monitoring setting due to high variance in the estimates. 

Finally, most earlier approaches have been examined only empirically in a limited number of settings and come with no theoretical guarantees. Alternatively, where such guarantees are given, they work only under strict conditions~\citep{garg:2022} or require extensive computational resources~\citep{chen:2021}. In contrast, CBPE requires only that the estimated model's confidence scores are calibrated. It has already been shown that under this \textit{calibration assumption} CBPE yields unbiased and consistent estimates for model accuracy~\citep{kivimaki:2025}. In this paper, we extend these theoretical guarantees to other classification metrics. Furthermore, CBPE allows the user to derive the full probability distribution of the estimated metric, which can then be used to quantify uncertainty in the estimated metrics in the form of valid confidence intervals.

The main contributions of this paper are the following:
\begin{enumerate}
    \item We present CBPE, a novel method for deriving estimates for different binary classification metrics using calibrated confidence scores. 
    \item Each estimated metric comes in the form of a full probability distribution, allowing the user to form valid confidence intervals for the estimates.
    \item For each metric, we show a shortcut to derive or approximate the expected value in situations where deriving the full distribution might be computationally too expensive. These shortcuts come with strong theoretical justifications.
\end{enumerate}

\section{Related Works}

Several different strategies for solving the unsupervised accuracy estimation problem have been suggested, such as training an ensemble of models and comparing the predictions of the models in the ensemble~\citep{baek:2022,chen:2021,jiang:2022}. Another approach has tried to leverage importance weighting~\citep{lu:2022} to adjust the known performance in a source distribution to some unknown target distribution~\citep{chen2021b}. Approaches leveraging distributional distances have also been suggested~\citep{deng:2023,lu:2023}. 

In this work, we focus on \textit{confidence-based} estimators. These estimators utilize confidence scores produced by a model to estimate its performance under a given target distribution. Confidence scores are commonly used in ML models' \textit{uncertainty estimation} as a numerical indicator of a system's certainty that its output is correct. Many types of models produce such scores innately~\citep{guo:2017, zadrozny:2001, zadrozny:2002}, and there are ways to attach confidence scores to predictions of other models as well~\citep{kivimaki:2023}. Typically, the score is expressed on the interval $[0, 1]$, but there are some exceptions to this~\citep{wei:2022}.

Several confidence-based methods have been suggested, each exploiting different signals from model outputs. Average Confidence (AC)~\citep{hendrycks:2017,kivimaki:2025} provides a simple baseline by using the average of model’s confidence scores as an estimate for accuracy. Difference of Confidences (DoC)~\citep{guillory:2021} builds on this by training a simple regressor to map the difference of average confidences between source and target distributions as an estimate for model performance under the target distribution. Similarly, Average Thresholded Confidence (ATC)~\citep{garg:2022} learns thresholds that transfer across domains, allowing accuracy estimation on unseen distributions without labels. Finally, Nuclear Norm~\citep{deng:2023} analyzes both the confidence and diversity of prediction distributions across unlabeled data, aiming to better characterize error patterns. Some of these methods have variants, where the uncertainty of the model is expressed in terms of prediction entropy instead of model confidence~\citep{garg:2022,guillory:2021}. 

All of the above estimators can only be used to estimate model accuracy. As already briefly mentioned in Section~\ref{sec:intro}, accuracy is a problematic metric in many cases. For example, with imbalanced data, high accuracy can be achieved trivially by always predicting the majority class. To the best of our knowledge, no confidence-based estimator that can estimate any other performance metric has been proposed prior to the method presented in this paper.

\section{Background}\label{sec:bg}

In this section, we present the necessary background information. We begin with calibration, then briefly discuss different types of distributional shifts.

\subsection{Confidence Calibration}\label{subsec:calibration}

Let us assume that the binary classification model $f$ under scrutiny generates a confidence score $S\in [0,1]$ in addition to its predictions $\hat{Y}=f(X)$, where the features $X$ and label $Y$ are sampled from a joint distribution $(X, Y)\sim p(\boldsymbol{x},y)$. Then, we have
\begin{definition}\label{def:calibration}
    Model $f$ is perfectly calibrated within distribution $p(\boldsymbol{x},y)$ iff.
    \begin{equation*}
        P_{p(\boldsymbol{x},y)}(Y=1 \mid S=s) = s \quad \forall s \in [0,1].
    \end{equation*}
\end{definition}
In effect, this definition means that when a model is calibrated, any of its confidence scores can be interpreted as a probability that the predicted instance belongs to the positive class. In our theoretical analyses of binary classifiers, we sometimes refer to \textit{confidence-consistency}, which we define as 
\begin{definition}\label{def:consistent}
    Model $f$ is confidence-consistent in $p(\boldsymbol{x}, y)$ iff. $\forall s~\in~[0,1]~ \exists a \in \{0, 1\}$: $P_{p(\boldsymbol{x},y)}(\hat{Y}=a \mid S=s)=1$. 
\end{definition}


This property ensures that the model $f$ does not make contradictory predictions for the same confidence score. Typical examples of confidence-consistent classifiers are probabilistic binary classifiers with predictions based on confidence thresholding. That is, $\hat{y}=1$, when $s\geq t$ and $\hat{y}=0$ otherwise for some threshold $t\in [0, 1]$. 

\subsection{Shifts in the Data Distribution}\label{subsec:shift}

When training a supervised ML model $f$, we typically use data from a \textit{source} distribution $p_s(\boldsymbol{x},y)$. However, once deployed, the model will encounter data from a potentially different \textit{target} distribution $p_t(\boldsymbol{x},y)$, a scenario known as \textit{Dataset shift}~\citep{moreno-torres:2012}. If labels from the target distribution are inaccessible, the performance of $f$ under $p_t(\boldsymbol{x},y)$ can only be estimated. Under these conditions, assumptions about the nature of the shift are necessary \cite{garg:2022}. Since $p(\boldsymbol{x},y)$ can be factorized as either $p(y|\boldsymbol{x})p(\boldsymbol{x})$ or $p(\boldsymbol{x}|y)p(y)$, the shift can be attributed to following components:
\begin{itemize}
    \item \textit{concept shift}: $p_s(y|\boldsymbol{x}) \neq p_t(y|\boldsymbol{x})$ and $p_s(\boldsymbol{x}) = p_t(\boldsymbol{x})$
    \item \textit{covariate shift}: $p_s(\boldsymbol{x}) \neq p_t(\boldsymbol{x})$ and $p_s(y|\boldsymbol{x}) = p_t(y|\boldsymbol{x})$
    \item \textit{label shift}: $p_s(y) \neq p_t(y)$ and $p_s(\boldsymbol{x}|y) = p_t(\boldsymbol{x}|y)$
\end{itemize}
Intuitively, concept shift means that the feature distribution remains unchanged, but the conditional distribution of labels changes. This is typical in adversarial settings, such as fraud detection, or in situations where model predictions affect user behavior, such as in recommender systems. In covariate shift, the opposite happens. This is typical in situations where the relation between features and labels rarely changes, such as when modelling diseases based on symptoms. Such a model might be trained on some (unrepresentative) subpopulation but applied to the whole population. Alternatively, during an epidemic, the prevalence of a disease in a fixed population might change over time, making it a case of label shift. 

\section{Methodology}\label{sec:CBPE}
Throughout this section, we are interested in monitoring a probabilistic binary classifier, which produces a prediction $\hat{Y}_i = f(X_i)$ and a calibrated confidence score $S_i = s(X_i)$ for each instance $X_i$. For simplicity, the monitored predictions are assumed to arrive in batches of fixed size $n$. Furthermore, we assume that we do not have access to actual labels $Y_i$ so we can not calculate any classification metrics directly, but have to rely on probabilistic estimates. We will denote the sets of indices for the positive and negative predictions as $I^+ = \{i:\hat{Y}_i=1\}$ and $I^- = \{i:\hat{Y}_i=0\}$, respectively.

\subsection{Estimating the Confusion Matrix}
Our approach begins by estimating the elements of the confusion matrix for a batch of $n$ predictions. Assume there are $n_{+}$ positive and $n_{-}$ negative predictions. Next, we will need to estimate how many of the positive and negative predictions actually belong to the positive and negative classes, respectively. 

Let us start with the positive predictions. Each positive prediction is either a \textit{true positive} $(TP)$ or a \textit{false positive} $(FP)$. Thus, it suffices to estimate the number of $TP$, since $|TP| + |FP| = n_{+}$. Since our model is assumed to be perfectly calibrated, we can interpret a confidence score of a given prediction as the probability that the predicted instance belongs to the positive class. Thus, we consider the number of $TP$ to be a (discrete) random variable $X_{TP}$, which follows a Poisson binomial distribution, where the confidence scores for the positive predictions $\{S_i: i\in I^+\}$ act as parameters. If we are interested in having a point estimate for the number of $TP$, we can use the expected value of the distribution, $\mathbb{E}[X_{TP}]=\sum_{i\in I^+} S_i$, as such\footnote{Note, that expectations such as this are conditional on the data $\{X_i\}_{i=1}^n$ and thus random variables until data is observed. However, we use $\mathbb{E}[X_{TP}]$ instead of more precise $\mathbb{E}[X_{TP} \mid \{X_i\}_{i=1}^n]$ for brevity.}. The number of $FP$ is then a derived random variable $X_{FP} = n_{+} - X_{TP}$. Also, since 
\begin{equation*}
    \mathbb{E}[X_{TP}] + \mathbb{E}[X_{FP}] = \mathbb{E}[X_{TP}+ X_{FP}] = \mathbb{E}[n_{+}] = n_{+},
\end{equation*}the expected value of $X_{FP}$ is 
\begin{equation*}
    \mathbb{E}[X_{FP}] = n_{+}-\mathbb{E}[X_{TP}] = n_{+} - \sum_{i\in I^+} S_i = \sum_{i\in I^+} (1-S_i).
\end{equation*}

Similarly, for negative predictions, the numbers of \textit{true negatives} $(TN)$ and \textit{false negatives} $(FN)$ can also be estimated using the confidence scores. As the calibrated confidence scores are probabilities that a given instance belongs to the positive class, the probability of a given instance $Y_i$ belonging to the negative class is the complement of its confidence score, formally, $P(Y_i=0) = 1-S_i$. Thus, the number of $TN$ is a random variable $X_{TN}$, which again follows a Poisson binomial distribution, where the complements $1 - S_i$ of the confidence scores $\{S_i: i \in I^-\}$ act as parameters. A point estimate for the number of $TN$ is then the expectation $\mathbb{E}[X_{TN}]=\sum_{i \in I^-} (1-S_i)$. Finally, the number of $FN$ is a random variable $X_{FN}$ derived from $X_{FN} = n_{-} - X_{TN}$, with the expected value $\mathbb{E}[X_{FN}] = \sum_{i \in I^-} S_i$ . 

The point estimates for the elements of the confusion matrix have been collected in Table~\ref{table:ec}.

\begin{table}[ht!]
\scriptsize
\centering
\caption{Point Estimates for the Elements of the Estimated Confusion Matrix}
\begin{tabular}{rr||c|c} 
& & \multicolumn{2}{c}{Predicted}\\
 & & Positive & Negative \\ 
\hline \hline
&&&\\
\parbox[t]{0mm}{
\multirow{2}{*}{\rotatebox[origin=c]{90}{Actual~~}}} & 
Positive & \(\displaystyle \mathbb{E}[X_{TP}] = \sum_{i \in I^+} S_i\) & \(\displaystyle \mathbb{E}[X_{FN}] = \sum_{i \in I^-} S_i\)  \\[4ex]
& Negative & \(\displaystyle \mathbb{E}[X_{FP}] = \sum_{i \in I^+} 1-S_i\) & \(\displaystyle \mathbb{E}[X_{TN}] = \sum_{i \in I^-} 1-S_i\) \\
&&&\\
\hline
\end{tabular}
\label{table:ec}
\end{table}

The goal of performance estimation in a monitoring setting is to detect persistent changes in model performance caused by changes in the target data distribution $p_t(\boldsymbol{x}, y)$. As we plan to use the estimated confusion matrix in this task, we would like to have some theoretical guarantees for the quality of our estimates. However, the values of the elements of the confusion matrix are expressed in absolute quantities, which makes them dependent on sample size and prone to sampling error. Thus, in order to provide the aforementioned guarantees on a distributional level, we need to move from absolute quantities to relative frequencies, which do not depend on sample size. 

Let the \textit{True Positive Frequency} of a model $f$ operating on $p_t(\boldsymbol{x}, y)$ mean the probability $P_{p_t(\boldsymbol{x},y)}(Y=1\mid f(X)=1)$ and denote it with $\operatorname{TPF}_t(f)$. Similarly, we can define frequencies of the other elements of the confusion matrix and denote them with $\operatorname{FPF}_t(f)$, $\operatorname{TNF}_t(f)$, and $\operatorname{FNF}_t(f)$. Assume we sample $n$ points from $p_t(\boldsymbol{x}, y)$ and collect predictions for them. Denote the means of confidence scores for the positive and negative predictions within the sample with $\bar{S}^+ = \frac{1}{n_{+}}\sum_{i \in I^+}S_i$ and $\bar{S}^- = \frac{1}{n_{-}}\sum_{i \in I^-}S_i$ respectively. Given this nomenclature, we can state the following theorem, for which the proof is given in Appendix~\ref{app:proofs}.  

\begin{theorem}\label{th:uc}
   Let $f$ be a perfectly calibrated and confidence-consistent model operating on some target distribution $p_t(\boldsymbol{x}, y)$. Then, $\bar{S}^+$, $\bar{S}^-$, $1-\bar{S}^+$, and $1-\bar{S}^-$ are unbiased and consistent estimators for $\operatorname{TPF}_t(f)$, $\operatorname{FPF}_t(f)$, $\operatorname{TNF}_t(f)$, and $\operatorname{FNF}_t(f)$, respectively.
\end{theorem}

The significance of this theorem is that it guarantees that the point estimates of the estimated confusion matrix derived from a finite sample reflect how the model operates over the whole underlying target distribution $p_t(\boldsymbol{x}, y)$. The only difference is that these guarantees are given in terms of relative frequencies, whereas the point estimates are given in absolute terms and depend on the sample size. These two are connected in a straightforward way. For example, 
$\mathbb{E}[X_{TP}]=\sum_{i \in I^+} S_i = n_{+}\bar{S}^+$.

\subsection{Estimating Classification Metrics}

In this paper, we devise algorithms for estimating four classification metrics used in binary classification: accuracy, precision, recall, and F$_1$. 
We treat these metrics as random variables and derive their full distribution. While the expressions for these derived random variables follow straightforwardly from the standard definitions, treating the estimated metrics as random variables allows us to capture the whole distribution of the estimated metric, which enables us to derive valid confidence intervals for the estimates. Furthermore, deriving the exact distributions using these expressions might require non-trivial calculations for some metrics. We will present algorithms for two such metrics, namely, recall and F$_1$.

First, we need to estimate the elements of the confusion matrix using the calibrated confidence scores from a given sample. In what follows, we assume that this has already been done. Similarly to the estimated elements of the confusion matrix, after the full distribution of the derived random variable is known, one can use the expected value of the distribution as a point estimate for the metric of interest.

\subsubsection{Accuracy} Since $X_{TP}+X_{FP}+X_{TN}+X_{FN}=n$ is fixed, accuracy can be regarded as the following derived random variable:
\begin{equation}
    X_{accuracy} = \frac{X_{TP}+X_{TN}}{n}.
\end{equation}
A naive way of forming the distribution of accuracy would be to first form the joint distribution of $X_{TP}$ and $X_{TN}$, which is straightforward since clearly $X_{TP} \ind X_{TN}$, so the joint is just the product of the marginal distributions. After that, one can iterate over the (pairwise) values of the joint and collect the sums and their respective probabilities. A simple shortcut is provided by realizing that 
\begin{equation}
    X_{accuracy} = \frac{X_{correct}}{n},
\end{equation}
where $X_{correct}$ can be derived in the following way: For all positive predictions, $S_i$ can directly be taken as the probability that the prediction was correct. On the other hand, for all negative predictions, one can interpret $1-S_i$ as the probability that the prediction was correct. Thus, one can map the confidence scores $S_i \rightarrow Z_i$ with
\begin{equation}
    Z_i=
    \begin{cases}
        S_i, & \hat{Y}_i = 1 \\
        1-S_i, & \hat{Y}_i = 0.
    \end{cases}
\end{equation}
Then, $X_{correct}$ follows a Poisson binomial distribution with $Z_i$ as parameters. Taking the expected value of this distribution yields the Average Confidence (AC)~\citep{guillory:2021, hendrycks:2017} estimator, which has already been shown to be an unbiased and consistent estimator of model accuracy under the calibration assumption~\citep{kivimaki:2025}.

\subsubsection{Precision} Similarly, since $X_{TP}+X_{FP}=n_{+}$ is fixed, the derived random variable capturing the distribution of precision is
\begin{equation}
    X_{precision} = \frac{X_{TP}}{X_{TP}+X_{FP}} = \frac{X_{TP}}{n_{+}}.
\end{equation}

\subsubsection{Recall}
Recall can be derived from the estimated confusion matrix with
\begin{equation}\label{eq:recall}
    X_{recall} = \frac{X_{TP}}{X_{TP}+X_{FN}}.
\end{equation}
As with accuracy, the joint distribution of $X_{TP}$ and $X_{FN}$ is the product of the marginal distributions since $X_{TP} \ind X_{FN}$. Thus, we can form the distribution of $X_{recall}$ by iterating over the joint $P(X_{TP}, X_{FN})$ and collecting all encountered values of the expression on the right side of the Equation~(\ref{eq:recall}) with their respective probabilities.
A small shortcut is to realize that $X_{recall} = 0$ whenever $X_{TP}=0$, and $X_{recall} = 1$ whenever $X_{FN}=0$, with the exception of both $X_{TP}$ and $X_{FN}$ being zero, which means that recall is undefined. In most practical situations, the probability of this event is negligible, but a user should be prepared to deal with it in one way or another. In our implementation, we set the value of the metric to be 0 in these cases for simplicity.

The exact details of how the distribution of recall is derived are given in Algorithm~\ref{alg:recall}. We assume that the input distributions are given in the form of associative arrays, where the values of the random variables act as keys (ranging from 0 to $n_{+}$ or $n_{-}$) and each value corresponding to a given key represents the probability of observing that key. On lines 1-3 we collect the maximum key value for both $X_{TP}$ and $X_{FN}$ and initialize an empty associative array for the distribution of $X_{recall}$. On line 4, we set the probability of recall being zero to be equal to the probability that the number of true positives is zero. On line 5 we set the probability of recall being one to be equal to the probability that the number of false negatives is zero minus the probability of true positives and false negatives both being zero. On lines 6-14 we iterate over all combinations $(i, j)$ of non-zero keys with $i$ from $X_{TP}$ and $j$ from $X_{FN}$, and check if the key $\frac{i}{i+j}$ already exists in the associative array describing the distribution of $X_{recall}$. If not, the probability of observing key $\frac{i}{i+j}$ will become the product of the probabilities $X_{TP}[i]$ and $X_{FN}[j]$. Otherwise, the probability of observing the key $\frac{i}{i+j}$ is increased by the amount of that product. 

The worst-case time complexity of the algorithm is $\mathcal{O}(n^2)$ since there are $n_{+}\cdot n_{-}$ iteration steps with $n_{+}+n_{-}=n$ and given a hash table implementation, the search and insertion operations can be performed in $\mathcal{O}(n)$ time, while all other operations are constant time\footnote{The Poisson binomial distributions for the confusion matrix elements can be constructed using an efficient algorithm based on the Fast Fourier Transform with a time complexity of $\mathcal{O}(n \log n)$~\citep{hong:2013}.}.

\begin{algorithm}[ht]
\caption{Derive Distribution of Recall}\label{alg:recall}
\begin{algorithmic}[1]
\Require $(X_{TP},X_{FN})$\quad(Distributions for $TP$ and $FN$)
\Ensure $X_{recall}$ \quad\quad\quad~~(Distribution of recall)
\State $n_{+} \gets |X_{TP}|-1$
\State $n_{-} \gets |X_{FN}|-1$
\State $X_{recall} \gets \{\}$ \quad\quad\quad (Empty associative array)
\State $X_{recall}[0] \gets X_{TP}[0]$
\State $X_{recall}[1] \gets X_{FN}[0] - (X_{TP}[0] \cdot X_{FN}[0]) $
\For{$i\leftarrow 1$ to $n_{+}$}
    \For{$j\leftarrow 1$ to $n_{-}$}
        \If{key $\frac{i}{i+j}$ already exists in $X_{recall}$}
            \State $X_{recall}\left[\frac{i}{i+j}\right] \gets X_{recall}\left[\frac{i}{i+j}\right] + X_{TP}[i] \cdot X_{FN}[j]$
        \Else
            \State $X_{recall}\left[\frac{i}{i+j}\right] \gets X_{TP}[i] \cdot X_{FN}[j]$
        \EndIf
    \EndFor
\EndFor
\end{algorithmic}
\end{algorithm}

\subsubsection{F1}
F$_1$ can be written as a derived random variable with the expression
\begin{equation}\label{eq:f1}
    X_{F_1} = \frac{2 \cdot X_{TP}}{2 \cdot X_{TP}+X_{FP}+X_{FN}} = \frac{2 \cdot X_{TP}}{ X_{TP}+X_{FN}+n_{+}},
\end{equation}
where $n_{+}$ is fixed. Similarly to recall, we can form the distribution $X_{F_1}$ by iterating over the joint distribution of $X_{TP}$ and $X_{FN}$ and collecting all encountered values of the expression on the right side of Equation (\ref{eq:f1}) with their respective probabilities. The details of how the distribution of F$_1$ is derived are given in Algorithm~\ref{alg:f1} with the same assumptions and notation as was used with Algorithm~\ref{alg:recall}. The logic in Algorithm~\ref{alg:f1} is completely analogous to that of Algorithm~\ref{alg:recall}. The only difference is in replacing the formula for calculating recall with the formula for calculating F$_1$. The worst-case time complexity of the algorithm is also identical to Algorithm~\ref{alg:recall}, namely $\mathcal{O}(n^2)$.

\begin{algorithm}[ht]
\caption{Derive Distribution of F$_1$}\label{alg:f1}
\begin{algorithmic}[1]
\Require $(X_{TP},X_{FN})$\quad(Distributions for $TP$ and $FN$)
\Ensure $X_{F_1}$ \quad\quad\quad\quad~~(Distribution of F$_1$)
\State $n_{+} \gets |X_{TP}|-1$
\State $n_{-} \gets |X_{FN}|-1$
\State $X_{F_1} \gets \{\}$; \quad\quad\quad\quad (Empty associative list)
\State $X_{F_1}[0] \gets X_{TP}[0]$
\For{$i\leftarrow 1$ to $n_{+}$}
    \For{$j\leftarrow 0$ to $n_{-}$}
        \If{key $\frac{2i}{i+j+n_{+}}$ already exists in $X_{F_1}$}
            \State $X_{F_1}\left[\frac{2i}{i+j+n_{+}}\right] \gets X_{F_1}\left[\frac{2i}{i+j+n_{+}}\right] + X_{TP}[i] \cdot X_{FN}[j]$
        \Else
            \State $X_{F_1}\left[\frac{2i}{i+j+n_{+}}\right] \gets X_{TP}[i] \cdot X_{FN}[j]$
        \EndIf
    \EndFor
\EndFor 
\end{algorithmic}
\end{algorithm}

\subsection{Shortcuts for Point Estimates}\label{sec:shortcuts}
Once the full distribution of a given metric is derived, one can use the expected value of the distribution as a point estimate.
However, with accuracy and precision, one does not need to form the full distributions of the metrics (or even estimate the confusion matrix) to derive the expectations directly from the confidence scores. We will now show how to apply a shortcut for the expectations of these metrics.

\subsubsection{Accuracy and Precision}
In the case of accuracy, we can calculate the expectation of accuracy directly by
\begin{equation*}
    \mathbb{E}[X_{accuracy}] = \mathbb{E}\left[\frac{X_{correct}}{n}\right] = \frac{1}{n} \sum_{i=1}^n Z_i,
\end{equation*}

Similarly, we can calculate the expectation of precision directly with
\begin{equation*}
    \mathbb{E}[X_{precision}] = \mathbb{E}\left[\frac{X_{TP}}{n_{+}}\right] = \frac{1}{n_{+}} \sum_{i\in I^{+}} S_i,
\end{equation*}

\subsubsection{Recall and F1}

With recall and F$_1$, and a sufficiently large monitoring window, highly accurate approximations of expectations can be calculated using the confidence scores. These approximations rely on the following lemma, which we will prove in the Appendix~\ref{app:proofs}.
\begin{lemma}\label{lemma:convergence}
    Let $X=X_1+...+X_{n_1}$ and $Y=Y_1+...+Y_{n_2}$ be random variables following Poisson binomial distributions with $X \ind Y$. Moreover, assume that there exist $\delta \in (0,\frac{1}{2}]$ and $N \in \mathbb{N}$ such that for all $n=n_1+n_2\geq N$, $\delta \leq \frac{n_1}{n} \leq 1-\delta$. Finally, assume that $\operatorname{Var}[X]\rightarrow\infty$ and $\operatorname{Var}[Y]\rightarrow\infty$, when $n\rightarrow\infty$ and $0\leq a \leq 1$. Then,
    \begin{equation}
        \mathbb{E}\left[\frac{X}{X+Y+an}\right] \rightarrow
        \frac{\mathbb{E}[X]}{\mathbb{E}[X]+\mathbb{E}[Y]+an},
    \end{equation}
    when $n\to\infty$.  Furthermore, the decay rate of the approximation error 
    is $\mathcal{O}\left(\frac{1}{\sqrt{n}}\right)$.
\end{lemma}

With recall, we can leverage Lemma~\ref{lemma:convergence} (with $a=0$) to approximate the expectation with
\begin{equation*}
    \mathbb{E}[X_{recall}] = \mathbb{E}\left[\frac{X_{TP}}{X_{TP}+X_{FN}}\right] \approx \frac{\mathbb{E}[X_{TP}]}{\mathbb{E}[X_{TP}] + \mathbb{E}[X_{FN}]} = 
    \frac{\sum\limits_{i\in I^{+}} S_i}{\sum\limits_{i\in I^{+}} S_i + \sum\limits_{i\in I^{-}} S_i} =
    \frac{\sum\limits_{i \in I^{+}} S_i}{\sum\limits_{i=1}^n S_i}.
\end{equation*}

Similarly, we can use Lemma~\ref{lemma:convergence} (with $a=\frac{n_{+}}{n}$) to approximate the expectation of F$_1$ as
\begin{equation*}
    \mathbb{E}[X_{F_1}] = \mathbb{E}\left[\frac{2 \cdot X_{TP}}{X_{TP}+X_{FN}+n_{+}}\right] \approx \frac{2 \cdot \mathbb{E}[X_{TP}]}{\mathbb{E}[X_{TP}] + \mathbb{E}[X_{FN}] + n_{+}} = \frac{2\cdot\sum\limits_{i\in I^{+}}S_i}{\sum\limits_{i=1}^n S_i + n_{+}}.
\end{equation*}

These simple formulas speed up calculations significantly when deriving point estimates for the performance metrics in question, which might be useful if the monitoring window contains a large number of samples. If the monitoring window size is small, the approximations for recall and F$_1$ might be too imprecise for practical purposes. In such computationally lightweight cases, using the exact values derived from the full distribution of the metric is preferable. We will return to this issue in Section~\ref{exp:shortcut}, where we explore the decay rates of the approximation errors empirically.



\section{Experiments}\label{experiments}
In this section, we conduct some experiments to back up our theoretical claims. We begin by analyzing the empirical convergence rate of the shortcut estimators presented in Section~\ref{sec:shortcuts}. Then, we examine the quality of the confidence intervals of our proposed estimator, and finally, move to experiments with real-life data.\footnote{We make the code available at \url{https://github.com/JuhaniK/
CBPE-experiments}} 

\subsection{Convergence of the Shortcut Estimators}\label{exp:shortcut}
\begin{figure}[ht!]
\centering
\includegraphics[width=0.67\columnwidth]{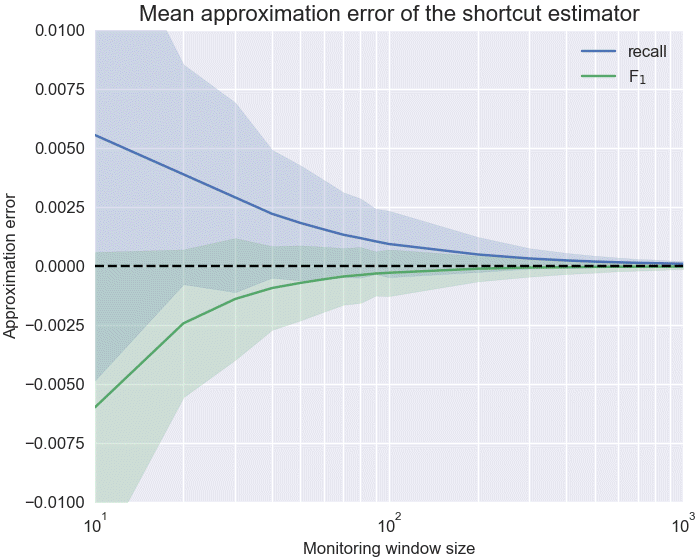}
\caption{Mean approximation error (the difference of true and approximated expectations) of the shortcut estimators in 10,000 trials for each monitoring window size from 10 to 1,000. The shaded area depicts one standard deviation.}
\label{fig:error}
\end{figure}

We used Lemma (\ref{lemma:convergence}) to give asymptotic guarantees for the approximations of the expectations of recall and F$_1$. Here, we demonstrate experimentally the rate of convergence in practice. For this, we created synthetic data by sampling confidence scores from different beta distributions, derived the full distributions for the metrics, and calculated the expected values. We then compared these values to the approximations derived using the shortcut formulas presented in Section~\ref{sec:shortcuts} and recorded the approximation error. We repeated this procedure 10,000 times for each monitoring window size, sampling from a different beta distribution, whose $\alpha$ and $\beta$ parameters had been (uniformly) randomly selected from the interval $[0.1, 10]$ in each trial. We used monitoring window sizes ranging from 10 to 1,000. The results are shown in Figure~\ref{fig:error}.

The figure shows that the approximation error decreases rapidly, with the mean absolute error already below 0.001 at window size 100. The deviation of the approximation error also goes to zero quite rapidly, which means that the shortcuts yield consistent estimates for the metrics in question. It is to be noted that this analysis does not depend in any way on whether the confidence scores are calibrated or not. Based on these findings, we recommend deriving the full distributions of the estimated metrics for small monitoring window sizes and using the shortcut estimators when the monitoring window size is big (500+, depending on the situation and user needs) and when one is not interested in the confidence intervals of the estimates.  

\subsection{Quality of the Confidence Intervals}\label{sec:CI}
In this experiment, we used the same synthetic data set as in the experiment described in Section~\ref{exp:shortcut}, but with the addition of using the sampled confidence scores to sample a corresponding label for each confidence score from a Bernoulli distribution using the confidence score as a parameter. This way of reverse sampling ensures that the simulated classifier is calibrated by construction although a small calibration error might persist due to sampling effects. After this, we again conducted 10,000 trials for each monitoring window size, ranging from 100 to 1,000, and recorded whether the actual value was within a predicted 95\% or 90\% confidence interval. Finally, we calculated the ratio of these successful trials over all trials, which should be close to 0.95 or 0.9, respectively. 

\begin{figure}[ht!]
\centering
\includegraphics[width=0.95\columnwidth]{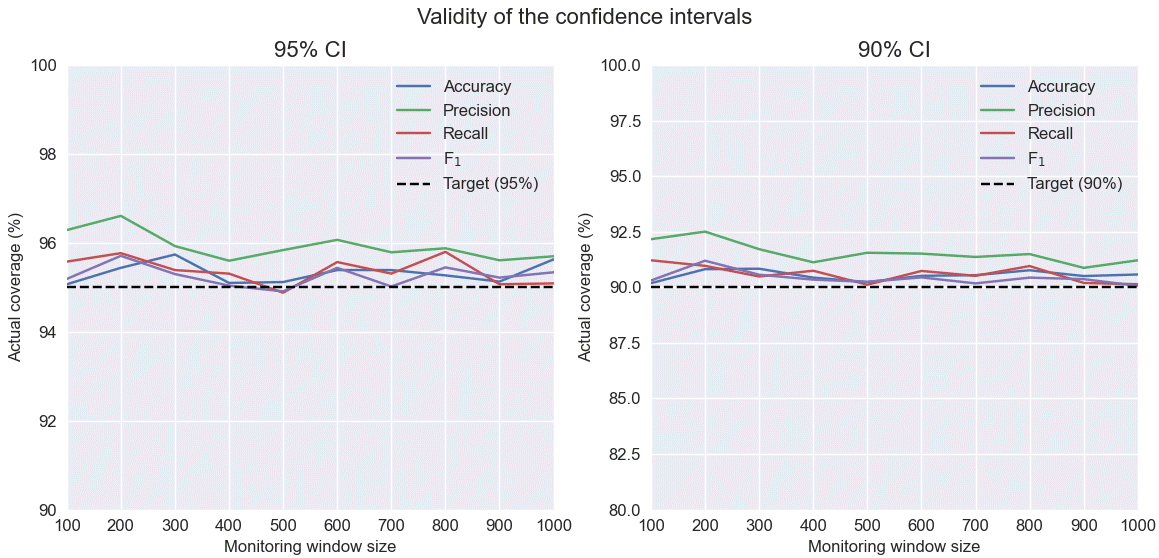}
\caption{The fraction of times the actual value of a metric was within the predicted confidence interval over 10,000 trials.}
\label{fig:ci}
\end{figure}

The results are shown in Figure~\ref{fig:ci}, which shows that the perceived ratios correspond closely to nominal targets. Furthermore, the deviation of the perceived ratios is on the positive side, making the confidence intervals slightly conservative, which is usually preferable.

\subsection{Tableshift Benchmark}\label{sec:benchmark}
For experiments with real-life data, we used datasets from the recently published TableShift benchmark~\citep{gardner:2024}. It provides an API to access 15 tabular datasets collected from various sources to test the robustness of binary classifiers under distribution shift. Here we use the benchmark outside of its original scope to see how well CBPE estimates different metrics within and outside the original data distribution. Unfortunately, the API is not well maintained, which made accessing most of the datasets somewhat difficult. For most datasets, we overcame these issues, but had to leave three datasets out of our experiments due to being unable to access them. In addition, four of the datasets require credentialed access, so these were also left out of our experiments. The remaining eight still present a heterogeneous collection of datasets\footnote{Full descriptions of these datasets are given at \url{https://tableshift.org/datasets.html}} from diverse sources and domains to offer relevant insight into the applicability of our proposed method. 

We trained an XGBoost model for each of the remaining eight datasets. We optimized the hyperparameters using Bayesian optimization over the search spaces defined in the original benchmark. The performance of the models was then estimated with CBPE for all four metrics of interest using unseen test sets from both in-distribution (ID) and out-of-distribution (OoD) splits. We ran 1,000 trials using a monitoring window size of 500. That is, for each trial, we sampled 500 random data points from the test sets. We then extracted the true values of the metrics and calculated the estimation error of the CBPE method. The results are presented in Table~\ref{tab:u_results_sample}, where for both splits we report class imbalance (the prevalence of the positive class) as "Bias", the \textit{Adaptive Expected Calibration Error}~\cite{nixon:2019} of each model as "ACE", the actual values for the four metrics of interest, and the estimation errors for the CBPE method as \textit{Mean Absolute Error} (MAE). We also provide error bounds for the estimation errors in the form of two standard deviations. All values are given as percentage units. 

\begin{sidewaystable}[p]
    \centering
    \caption{Mean Absolute Estimation Errors for Different Metrics with TableShift Datasets (\%)}
    \begin{tabular}{l|c|c|c|c|c|c|c|c|c|c|c}
    \toprule
    \multicolumn{4}{c}{} & \multicolumn{2}{|c}{Accuracy} & \multicolumn{2}{|c}{Precision} & \multicolumn{2}{|c}{Recall} & \multicolumn{2}{|c}{F$_1$} \\
    \midrule
        Dataset & Split & Bias & ACE & True & MAE & True & MAE & True & MAE & True & MAE \\ \hline
        \multirow{2}{*}{ASSISTments} & ID & 69.5 & 0.26 & 94.2 & $0.01\pm1.0$ & 93.3 & $0.06\pm1.2$ & 98.7 & $0.04\pm0.7$ & 96.0 & $0.01\pm0.7$ \\ 
        & OoD & 43.7 & 31.9 & 58.6 & $31.6\pm0.9$ & 51.3 & $37.5\pm1.0$ & 99.4 & $0.14\pm0.5$ & 67.7 & $26.2\pm0.6$ \\ \hline
        College & ID & 12.7 & 0.48 & 95.7 & $0.03\pm0.9$ & 87.2 & $1.40\pm4.2$ & 77.1 & $0.06\pm6.9$ & 81.9 & $0.63\pm4.5$ \\ 
        Scorecard & OoD & 31.1 & 3.52 & 87.9 & $0.05\pm1.2$ & 86.8 & $2.84\pm2.7$ & 71.9 & $1.08\pm4.9$ & 78.6 & $0.59\pm3.2$ \\ \hline
        \multirow{2}{*}{Diabetes} & ID & 12.7 & 0.43 & 87.7 & $0.19\pm1.1$ & 58.1 & $0.62\pm3.2$ & 12.3 & $0.08\pm6.0$ & 20.3 & $0.10\pm8.2$ \\ 
        & OoD & 17.4 & 3.87 & 83.3 & $3.63\pm1.2$ & 60.5 & $2.71\pm2.8$ & 12.2 & $2.67\pm6.3$ & 20.3 & $3.19\pm8.0$ \\ \hline
        \multirow{2}{*}{Food Stamps} & ID & 19.1 & 0.51 & 84.9 & $0.18\pm1.3$ & 66.4 & $0.69\pm2.9$ & 42.6 & $0.39\pm7.2$ & 51.9 & $0.41\pm5.5$ \\ 
        & OoD & 22.0 & 1.96 & 82.6 & $1.60\pm1.3$ & 64.4 & $3.11\pm2.6$ & 46.2 & $3.50\pm6.7$ & 53.8 & $3.38\pm4.7$ \\ \hline
        Hospital & ID & 41.6 & 3.02 & 66.6 & $0.39\pm1.0$ & 62.7 & $0.15\pm1.5$ & 48.9 & $0.47\pm5.1$ & 54.9 & $0.28\pm3.4$ \\ 
        Readmission & OoD & 49.4 & 4.05 & 62.5 & $1.94\pm0.9$ & 65.5 & $2.80\pm1.4$ & 50.8 & $2.18\pm4.7$ & 57.2 & $0.18\pm2.9$ \\ \hline
        \multirow{2}{*}{Hypertension} & ID & 40.2 & 1.23 & 67.0 & $0.41\pm1.0$ & 61.8 & $0.84\pm1.4$ & 46.7 & $0.57\pm4.9$ & 53.2 & $0.72\pm3.3$ \\ 
        & OoD & 58.4 & 14.8 & 60.4 & $6.21\pm0.9$ & 75.3 & $11.4\pm1.3$ & 48.0 & $6.38\pm4.7$ & 58.6 & $0.08\pm2.9$ \\ \hline
        \multirow{2}{*}{Income} & ID & 32.1 & 0.62 & 83.1 & $0.29\pm1.3$ & 75.3 & $0.71\pm2.2$ & 70.2 & $0.21\pm4.7$ & 72.7 & $0.42\pm2.8$ \\ 
        & OoD & 39.8 & 5.71 & 81.3 & $1.69\pm1.4$ & 82.1 & $6.13\pm2.1$ & 67.9 & $5.46\pm4.3$ & 74.3 & $0.30\pm2.6$ \\\hline
        \multirow{2}{*}{Unemployment} & ID & 03.4 & 0.10 & 97.3 & $0.04\pm0.7$ & 73.2 & $0.51\pm17$ & 32.5 & $1.20\pm18$ & 45.0 & $1.99\pm19$ \\ 
        & OoD & 05.2 & 0.14 & 96.2 & $0.14\pm0.9$ & 77.2 & $0.41\pm12$ & 36.8 & $0.74\pm15$ & 49.8 & $0.29\pm15$ \\ 
        \bottomrule
    \end{tabular}
    \label{tab:u_results_sample}
\end{sidewaystable}

The results show that CBPE results in low estimation errors when the calibration error is relatively small and in some cases even when the calibration error is rather large, as is the case with Hospital Readmission data. It is worthwhile to notice that all of these models are optimized for accuracy, and their performance with respect to other metrics might be lacking. This is most notable with the Diabetes data. This shows that CBPE can perform well for a wide range of true metric values. In terms of accuracy, the performance of all models drops significantly with the shift. This is most dramatic for the model trained with ASSISTments data, which also becomes markedly uncalibrated after the shift. Finally, the performance of the models trained for the Hospital Readmission and Hypertension datasets is lacking to begin with. Both are only slightly better than random guessing, indicating that the models struggle to find generalizable patterns from the training sets. The Hospital Readmission model is not calibrated even for the ID data, but its calibration error does not increase dramatically with the shift, whereas the Hypertension model is roughly calibrated in the ID split, but highly uncalibrated in the OoD split. 

The deviation in the estimation errors is relatively small for accuracy and precision, and somewhat higher for recall and F$_1$, except for the model trained with the Unemployment data, which has notably high variance in all estimation errors except for accuracy. The correlations between calibration errors and estimation errors were very strong for both accuracy and precision ($\rho = 0.960$ and $\rho = 0.983$, respectively), but interestingly less so for recall and F$_1$ ($\rho = 0.199$ and $\rho = 0.873$, respectively). 

\section{Discussion}\label{sec:discussion}
It is not immediately obvious how the methods for unsupervised accuracy estimation referenced in the Introduction could be adapted to estimate any other classification metric besides accuracy. As such, CBPE sets the baseline for unsupervised estimation of all binary classification metrics, which can be derived from the confusion matrix. Although CBPE comes with relatively strong theoretical guarantees, it also has its limitations. First and foremost, the theoretical guarantees hinge on the assumption of perfect calibration, which is not achievable in any real-life scenario. However, our experiments show that in most cases, CBPE yields consistent estimates with relatively small estimation errors, as the calibration errors remain small as well. This reduces the problem of performance estimation to a calibration error minimization problem.

Confidence-based performance estimators are known to struggle under concept shift, since even when $p_s(\boldsymbol{x}, y)$ and $p_t(\boldsymbol{x})$ are fully known, the problem of estimating model performance in this setting is fundamentally unidentifiable~\citep{garg:2022} and no guarantees for estimation quality can be given. This is because concept shift breaks the stochastic dependency between the features and labels, rendering the confidence scores of a model obsolete in most cases. Also, since calibration error typically increases with covariate shift~\citep{ovadia:2019} and label shift~\citep{popordanoska:2023}, methods for remaining calibrated under these shifts are needed. Unfortunately, no dedicated benchmarks for covariate shift and label shift scenarios in binary classification with tabular data currently exist. These issues present important topics for future research.

In this work, we used the TableShift benchmark as a proxy, despite the unspecified or unknown shift types for each dataset. Although the benchmark's authors have made an effort to quantify the amount of shift with regard to each component (covariate, concept, and label shift) for each dataset, they use ad hoc metrics instead of principled ones. In fact, the metrics used for covariate shift and concept shift in their work show almost linear dependency~\citep{gardner:2024}, hinting they might be measuring the same phenomenon rather than orthogonal components. The authors correctly point out that measuring the amount of shift is not possible from a finite sample~\citep{gardner:2024}. Indeed, no principled metrics exist for even approximating the shift from a finite sample in any of the components. Quantifying label shift is perhaps the easiest of the three, and propositions on how to quantify it have recently been made~\cite{sun:2025}. However, shift quantification remains an open research problem.

\section{Conclusion}

We presented CBPE, a family of confidence-based performance estimators for model monitoring, which require no access to ground truth labels.  Contrary to other confidence-based estimators, which can only target model accuracy, CBPE can estimate a binary classifier's performance with any metric derivable from the confusion matrix, making CBPE more suitable for monitoring classifiers under class imbalance or when some error type is more costly. Also, contrary to other estimators, CBPE quantifies uncertainty in the estimates by yielding a full probability distribution for the estimated metric. We presented theoretical guarantees of its quality under perfect calibration and showed experimentally that its estimation error remains feasible under small calibration error. We hope our work lays the groundwork for future research on unsupervised performance estimation.

\backmatter

\bmhead{Acknowledgements}

This work was partly funded by local authorities (“Business Finland”) under grant agreements 20219 IML4E and 23004 ELFMo of the ITEA4 programme.

\newpage
\begin{appendices}

\section{Review of Classification Metrics}\label{sec:metrics}

In binary classification, prediction-label pairs come in four types: A \textit{True Positive} $(TP)$ instance was predicted to belong to the positive class and this was the case. A \textit{False Positive} $(FP)$, was predicted to be positive but this was not the case. By replacing the positive with negative in the above, we can similarly define both a \textit{True Negative} $(TN)$ and a \textit{False Negative} $(FN)$ instances. A common practice is to present the numbers of each type for a given classifier and a given dataset in a Confusion Matrix. 

\begin{table}[ht!]
\small
\centering
\caption{Confusion Matrix}
\begin{tabular}{rr||c|c} 
& & \multicolumn{2}{c}{Predicted}\\
 & & Positive & Negative \\ 
\hline \hline
&&&\\
\parbox[t]{0mm}{
\multirow{2}{*}{\rotatebox[origin=c]{90}{Actual~~}}} & 
Positive & $TP$ & $FN$  \\[2ex]
& Negative & $FP$ & $TN$ \\
&&&\\
\hline
\end{tabular}
\label{table:confusion}
\end{table}

Once the confusion matrix has been formed, many different classification metrics can be derived from it. In this work, we focus on four of these, namely \textit{Accuracy}, \textit{Precision}, \textit{Recall}, and \textit{F}$_1$, since they are perhaps the most commonly used ones. Using the elements of the confusion matrix, they can be defined as follows:
\begin{align}
    \operatorname{Accuracy} &= \frac{TP+TN}{TP+FP+FN+TN} \\
    \operatorname{Precision} &= \frac{TP}{TP+FP} \\
    \operatorname{Recall} &= \frac{TP}{TP+FN} \\
    \operatorname{F_1} &= \frac{2}{\text{recall}^{-1}+\text{precision}^{-1}} = 2\frac{\text{precision}\cdot\text{recall}}{\text{precision}+\text{recall}} = \frac{2TP}{2TP+FP+FN} \label{eq:def:f1}
\end{align}

Accuracy measures the fraction of correct predictions, whether positive or negative. Precision, sometimes also referred to as \textit{positive predictive value}, measures the fraction of correct predictions out of all positive predictions. Recall, also known as \textit{sensitivity}, measures which fraction out of all instances belonging to the positive class were predicted as such. Finally, F$_1$ is the harmonic mean of Precision and Recall, which tries to balance the two. In all cases, the metric is measured on the interval $[0,1]$ and a higher number indicates better performance. Although we have chosen these four commonly used metrics under scrutiny, the approach we take in Section~\ref{sec:CBPE} could be applied to any metric, whose value can be calculated using the elements of the confusion matrix.

\section{Proofs}\label{app:proofs}
In the proofs to follow, we use the word "model" to refer exclusively to binary classifiers. We start with the proof of Theorem~\ref{th:uc} along with some complementary results. Then, we move on to prove Lemma~\ref{lemma:convergence}. 

\subsection{Proof of Theorem~\ref{th:uc}}
For this proof, we require an altered definition of calibration, where the probability of observing a positive label is also conditioned on the predicted label. We call this \textit{class-conditional calibration} with the following formal definition.

\begin{definition}\label{def:cc}
    Model $f$ is perfectly class-conditionally calibrated within $p_t(\boldsymbol{x},y)$ iff.
    \begin{equation*}
        P_t(Y=1 \mid \hat{Y}=a \cap S=s) = s \quad \forall (a, s) \in \operatorname{supp}(\hat{Y}, S).
    \end{equation*}
\end{definition}

One might be interested in how this concept relates to Definition~\ref{def:calibration}. It turns out that if we assume the model $f$ to be confidence-consistent (Definition~\ref{def:consistent}), then Definition~\ref{def:calibration} is a sufficient condition of Definition~\ref{def:cc}. We prove this in the following theorem.

\begin{theorem}\label{th:cc}
    Let model $f$ be perfectly calibrated and confidence-consistent. Then $f$ is also perfectly class-conditionally calibrated.
\end{theorem}
\begin{proof}
    Starting directly from Definition~\ref{def:calibration}, we get
    \begin{align*}
        P_t(Y=1 \mid S=s) &= s \\
        \sum_{c=0}^1 P_t(Y=1 \cap \hat{Y}=c \mid S=s) &= s \\
        \sum_{c=0}^1 P_t(Y=1 \mid \hat{Y}=c \cap S=s)P(\hat{Y}=c \mid S=s) &= s \\
        P_t(Y=1 \mid \hat{Y}=a \cap S=s) &= s,
    \end{align*}
    where the final step is validated by the fact that $f$ was assumed to be confidence-consistent, and hence for each $s\in [0, 1]$ there exists $a \in \{0, 1\}$ such that $P(\hat{Y}=a \mid S=s) = 1$, which guarantees that the final equation holds for all pairs $(a,s)$ within the support of $(\hat{Y}, S)$.
\end{proof}

Let us next consider a model $f$ operating on some target distribution $p_t(\boldsymbol{x}, y)$.
Suppose $p_t(s)$ denotes the density function of the distribution of confidence scores induced by model $f$ operating on $p_t(\boldsymbol{x}, y)$. 

\begin{lemma}\label{lemma:unbiased}
    Let $(X, Y)$ be an instance drawn from a target distribution $p_t(\boldsymbol{x}, y)$ and let $\hat{Y}$ be the corresponding prediction made with confidence $S$ by a perfectly calibrated and confidence-consistent model $f$. Then,
    \begin{equation*}
        \mathbb{E}_t[S\mid \hat{Y}=1] = P_t(Y=1 \mid \hat{Y}=1).
    \end{equation*}
\end{lemma}
\begin{proof}
    Consider the probability that the actual label is positive given that it was predicted to be positive with some confidence score $S=s$, and apply the Bayes rule to get
    \begin{align*}
        P_t(Y=1 \mid \hat{Y}=1 \cap S=s) &= \frac{p_t(\hat{Y}=1 \cap S=s \mid Y=1) P_t(Y=1)}{p_t(\hat{Y}=1 \cap S=s)} \\
        &= \frac{p_t(s \mid Y=1, \hat{Y}=1) P_t(\hat{Y}=1 \mid Y=1) P_t(Y=1)}{p_t(s \mid \hat{Y}=1) P_t(\hat{Y}=1)}.
    \end{align*}
    Since our assumptions guarantee the model to be perfectly class-conditionally calibrated, we can use Definition~\ref{def:cc} and substitute the left side to get
    \begin{align*}
        s &= \frac{p_t(s \mid Y=1, \hat{Y}=1) P_t(\hat{Y}=1 \mid Y=1) P_t(Y=1)}{p_t(s \mid \hat{Y}=1) P_t(\hat{Y}=1)} \\
        P_t(\hat{Y}=1)p_t(s \mid \hat{Y}=1)s &= p_t(s \mid Y=1, \hat{Y}=1) P_t(\hat{Y}=1 \mid Y=1) P_t(Y=1) \\        
        \int_0^1 P_t(\hat{Y}=1)p_t(s \mid \hat{Y}=1)s~ds &= \int_0^1 p_t(s \mid Y=1, \hat{Y}=1) P_t(\hat{Y}=1 \mid Y=1) P_t(Y=1)~ds \\
        P_t(\hat{Y}=1) \int_0^1 p_t(s \mid \hat{Y}=1)s~ds &= P_t(\hat{Y}=1 \mid Y=1)P_t(Y=1) \int_0^1 p_t(s \mid Y=1, \hat{Y}=1) ~ds \\
        P_t(\hat{Y}=1) \int_0^1 p_t(s \mid \hat{Y}=1)s~ds &= P_t(Y=1 \mid \hat{Y}=1)P_t(\hat{Y}=1) \int_0^1 p_t(s \mid Y=1, \hat{Y}=1) ~ds \\
        \int_0^1 p_t(s \mid \hat{Y}=1)s~ds &= P_t(Y=1 \mid \hat{Y}=1) \int_0^1 p_t(s \mid Y=1, \hat{Y}=1)~ds \\ 
        \mathbb{E}_t[S\mid \hat{Y}=1] &= P_t(Y=1 \mid \hat{Y}=1)
    \end{align*}
\end{proof}

\begin{corollary}\label{corollary}
    Let $(X, Y)$ be an instance drawn from a target distribution $p_t(\boldsymbol{x}, y)$ and let $\hat{Y}$ be the corresponding prediction made with confidence $S$ by a perfectly calibrated and confidence-consistent model $f$. Then, the following are true
    \begin{enumerate}
        \item $\mathbb{E}_t[S\mid \hat{Y}=0] = P_t(Y=1 \mid \hat{Y}=0)$
        \item $\mathbb{E}_t[1-S\mid \hat{Y}=1] = P_t(Y=0 \mid \hat{Y}=1)$
        \item $\mathbb{E}_t[1-S\mid \hat{Y}=0] = P_t(Y=0 \mid \hat{Y}=0)$
    \end{enumerate}
\end{corollary}
\begin{proof}
    Statement 1 follows directly from Lemma~\ref{lemma:unbiased} by substituting all instances of the expression $\hat{Y}=1$ with $\hat{Y}=0$ in the proof. Statement 2 also follows from Lemma~\ref{lemma:unbiased} by applying the complement rule as follows.
    \begin{align*}
        \mathbb{E}_t[S\mid \hat{Y}=1] &= P_t(Y=1 \mid \hat{Y}=1) \\
        1-\mathbb{E}_t[S\mid \hat{Y}=1] &= 1-P_t(Y=1 \mid \hat{Y}=1) \\
        \mathbb{E}_t[1-S\mid \hat{Y}=1] &= P_t(Y=0 \mid \hat{Y}=1) \\
    \end{align*}
    Similarly, Statement 3 follows from Statement 1 by applying the complement rule as above.
\end{proof}

We are now ready to prove Theorem~\ref{th:uc}. We split the proof into two parts. In the first part, we show the unbiasedness of the estimator, and in the second part, we show the consistency. Again, we sample $n$ points from $p_t(\boldsymbol{x}, y)$ and collect predictions for them.  
Assume that there are $n_{+}$ positive and $n_{-}$ negative predictions within the $n$ predictions. Denote the zero vector and a vector of all ones of size $m$ with $\bar{0}_m$ and $\bar{1}_m$ respectively. Let us start by proving the unbiasedness of our estimators in the following theorem.

\begin{theorem}\label{th:unbiased}
   Let $f$ be a perfectly calibrated and confidence-consistent model operating on some target distribution $p_t(\boldsymbol{x}, y)$. Then, the following are true
   \begin{enumerate}
        \item $\mathbb{E}_t\left[\bar{S}^+\mid \hat{Y}=\bar{1}_{n_{+}}\right] = \operatorname{TPF}_t(f)$
        \item $\mathbb{E}_t\left[\bar{S}^-\mid \hat{Y}=\bar{0}_{n_{-}}\right] = \operatorname{FNF}_t(f)$
        \item $\mathbb{E}_t\left[1-\bar{S}^+\mid \hat{Y}=\bar{1}_{n_{+}}\right] = \operatorname{FPF}_t(f)$
        \item $\mathbb{E}_t\left[1-\bar{S}^-\mid \hat{Y}=\bar{0}_{n_{-}}\right] = \operatorname{TNF}_t(f)$
    \end{enumerate}
\end{theorem}
\begin{proof}
    We only prove Statement 1, since by leveraging Corollary~\ref{corollary}, the structure of the proof is identical for all the other statements. Using linearity of expectation and Lemma~\ref{lemma:unbiased}, we have
    \begin{align*}
        \mathbb{E}_t\left[\bar{S}^+\mid \hat{Y}=\bar{1}_{n_{+}}\right] &= \mathbb{E}_t\left[\frac{1}{n_{+}}\sum_{i\in \mathrm{I}_+} S_i~\Bigg|~\hat{Y}=\bar{1}_{n_{+}}\right] \\
        &= \frac{1}{n_{+}}\sum_{i\in \mathrm{I}_+} \mathbb{E}_t[S_i \mid \hat{Y}_i=1] \\
        &= \frac{1}{n_{+}}\sum_{i\in \mathrm{I}_+} P_t(Y_i=1 \mid \hat{Y}_i=1) \\
        &= P_t(Y_i=1 \mid \hat{Y}_i=1) \\
        &= \operatorname{TPF}_t(\boldsymbol{\mathrm{f}})
    \end{align*}
\end{proof}

Next, we will show that the estimators are also consistent. By Chebyshev's inequality, it suffices to show that the variances of our estimators tend to zero when $n_{+}$ and $n_{-}$ go to infinity.

\begin{theorem}\label{th:consistent}
   Let $f$ be a perfectly calibrated and confidence-consistent model. Then, the following are true
   \begin{enumerate}
        \item $\underset{n_{+}\rightarrow\infty}{\lim}\operatorname{Var}\left[\bar{S}^+\right] = 0$
        \item $\underset{n_{-}\rightarrow\infty}{\lim}\operatorname{Var}\left[\bar{S}^-\right] = 0$
        \item $\underset{n_{+}\rightarrow\infty}{\lim}\operatorname{Var}\left[1-\bar{S}^+\right] = 0$
        \item $\underset{n_{-}\rightarrow\infty}{\lim}\operatorname{Var}\left[1-\bar{S}^-\right] = 0$
    \end{enumerate}
\end{theorem}
\begin{proof}
    Again, we show only the proof for Statement 1. The proofs for all other statements are completely analogous. The variance of $\bar{S}^+$ is
    \begin{equation*}
        \operatorname{Var}[\bar{S}^+] = \operatorname{Var}\left[\frac{1}{n_{+}}\sum_{i \in I^+} S_i\right] = \frac{1}{n_{+}^2}\sum_{i \in I^+}\operatorname{Var}\left[S_i\right] = \frac{1}{n_{+}^2}\sum_{i \in I^+}\left(\mathbb{E}[S_i^2]-\mathbb{E}[S_i]^2\right).
    \end{equation*}
    Since all $S_i$ take values from the interval $[0,1]$, it is trivial that $\mathbb{E}[S_i^2] \leq \mathbb{E}[S_i]$ for all $i$. Since clearly also $\mu := \mathbb{E}[S_i]\in [0,1]$, it is straightforward to see that the expression $\mu(1-\mu)$ is maximized when $\mu=\frac{1}{2}$, yielding a maximum of $\frac{1}{4}$. Using these insights, we can derive an upper bound for the variance of the estimator as follows.
    \begin{equation*}
        \operatorname{Var}[\bar{S}^+] \leq \frac{1}{n_{+}^2}\sum_{i \in I^+}\left(\mathbb{E}[S_i]-\mathbb{E}[S_i]^2\right) 
        = \frac{1}{n_{+}^2}\sum_{i \in I^+}\mu(1-\mu) 
        \leq \frac{1}{n_{+}^2}\sum_{i \in I^+}\frac{1}{4} 
        = \frac{1}{4n_{+}}.
    \end{equation*}
    Since $\underset{n_{+}\rightarrow\infty}{\lim} \frac{1}{4n_{+}} = 0$ and $0 \leq \operatorname{Var}[\bar{S}^+] \leq \frac{1}{4n_{+}}$, it follows that $\underset{n_{+}\rightarrow\infty}{\lim} \operatorname{Var}[\bar{S}^+] = 0$.
\end{proof}

Together, the proofs of Theorems~\ref{th:unbiased} and~\ref{th:consistent} constitute the proof of Theorem~\ref{th:uc}.

\subsection{Proof of Lemma~\ref{lemma:convergence}}

We restate the lemma here

\begin{replemma}{lemma:convergence}
    Let $X=X_1+...+X_{n_1}$ and $Y=Y_1+...+Y_{n_2}$ be random variables following Poisson binomial distributions with $X \ind Y$. Moreover, assume that there exist $\delta \in (0,\frac{1}{2}]$ and $N \in \mathbb{N}$ such that for all $n=n_1+n_2\geq N$, $\delta \leq \frac{n_1}{n} \leq 1-\delta$. Finally, assume that $\operatorname{Var}[X]\rightarrow\infty$ and $\operatorname{Var}[Y]\rightarrow\infty$, when $n\rightarrow\infty$ and $0\leq a \leq 1$. Then,
    \begin{equation}
        \mathbb{E}\left[\frac{X}{X+Y+an}\right] \rightarrow
        \frac{\mathbb{E}[X]}{\mathbb{E}[X]+\mathbb{E}[Y]+an},
    \end{equation}
    when $n\to\infty$. Furthermore, the decay rate of the approximation error 
     is $\mathcal{O}\left(\frac{1}{\sqrt{n}}\right)$.
\end{replemma}
\begin{proof}
    Consider the Taylor expansion of the function $f(X,Y)=\frac{X}{X+Y+an}$ at the point $(\mathbb{E}[X], \mathbb{E}[Y])$. Such an expansion is guaranteed to exist since $f$ is a smooth function. Trivially,
    \begin{equation*}
        f(\mathbb{E}[X], \mathbb{E}[Y]) = \frac{\mathbb{E}[X]}{\mathbb{E}[X]+\mathbb{E}[Y]+an}.
    \end{equation*}
    The (first-order) partial derivatives are 
    \begin{align*}
        \frac{\partial{f}}{\partial{X}} &= \frac{Y+an}{(X+Y+an)^2}, \\
        \frac{\partial{f}}{\partial{Y}} &= -\frac{X}{(X+Y+an)^2}.
    \end{align*}
    Thus, we can write the Taylor expansion as
    \begin{equation*}
        f(X,Y) = \frac{\mathbb{E}[X]}{\mathbb{E}[X]+\mathbb{E}[Y]+an} + \frac{(X-\mathbb{E}[X])(\mathbb{E}[Y]+an)}{(\mathbb{E}[X]+\mathbb{E}[Y]+an)^2} - \frac{(Y-\mathbb{E}[Y])\mathbb{E}[X]}{(\mathbb{E}[X]+\mathbb{E}[Y]+an)^2} + \epsilon(X,Y),
    \end{equation*}
    where $\epsilon(X,Y)$ holds the second and higher-order terms. 
    
    Since the maximum variance of any Bernoulli trial is $\frac{1}{4}$, it is easy to verify that Lindeberg's condition holds for both $X$ and $Y$. Thus, the Central Limit Theorem states that both $X-\mathbb{E}[X]$ and $Y-\mathbb{E}[Y]$ converge in distribution to a normal distribution, more precisely, 
    \begin{align*}
        \frac{X-\mathbb{E}[X]}{\sqrt{\operatorname{Var}(X)}} &= \frac{X-\mathbb{E}[X]}{\sqrt{n_1\sigma_{\bar{X}_i}^2}} = \frac{1}{\sqrt{n_1}}\cdot\frac{X-\mathbb{E}[X]}{\sigma_{\bar{X}_i}} \xrightarrow{~D~} \mathcal{N}(0,1), \\
        \frac{Y-\mathbb{E}[Y]}{\sqrt{\operatorname{Var}(Y)}} &= \frac{Y-\mathbb{E}[Y]}{\sqrt{n_2\sigma_{\bar{Y}_j}^2}} = \frac{1}{\sqrt{n_2}}\cdot\frac{Y-\mathbb{E}[Y]}{\sigma_{\bar{Y}_j}} \xrightarrow{~D~} \mathcal{N}(0,1),
    \end{align*}
    where $\sigma_{\bar{X}_i}^2$ and $\sigma_{\bar{Y}_j}^2$ are the average variances of $X_i$ and $Y_j$ respectively. Since $X_i$ and $Y_j$ are Bernoulli variables, we have the bounds $0<\sigma_{\bar{X}_i}\leq\frac{1}{2}$ and $0<\sigma_{\bar{Y}_j}\leq\frac{1}{2}$. From this, it necessarily follows that $X-\mathbb{E}[X]=\mathcal{O}_p(\sqrt{n_1})$ and $Y-\mathbb{E}[Y]=\mathcal{O}_p(\sqrt{n_2})$. Analyzing the growth rate of the other components in the expansion, we clearly have $\mathbb{E}[Y]+an=\mathcal{O}_p(n)$, $\mathbb{E}[X]=\mathcal{O}_p(n_1)$, and $(\mathbb{E}[X]+\mathbb{E}[Y]+an)^2=\mathcal{O}_p(n^2)$. Thus, we see that the first-order terms in the Taylor expansion converge (in probability) to zero with decay rates of
    \begin{align*}
        \frac{(X-\mathbb{E}[X])(\mathbb{E}[Y]+an)}{(\mathbb{E}[X]+\mathbb{E}[Y]+an)^2} &= \mathcal{O}_p\left(\frac{\sqrt{n_1}\cdot n}{n^2}\right) = \mathcal{O}_p\left(\frac{1}{\sqrt{n}}\right), \\
        - \frac{(Y-\mathbb{E}[Y])\mathbb{E}[X]}{(\mathbb{E}[X]+\mathbb{E}[Y]+an)^2} &= \mathcal{O}_p\left(\frac{\sqrt{n_2}\cdot n_1}{n^2}\right) = \mathcal{O}_p\left(\frac{1}{\sqrt{n}}\right).
    \end{align*}
    It is a straightforward (albeit tedious) exercise to show that the higher-order terms contained in $\epsilon(X, Y)$ also converge to zero, but with an even faster rate of decay (in fact, for the $k$th order terms, the decay rate is $\mathcal{O}_p\left(n^{-\frac{k}{2}}\right)$), which means that the decay rate of the first-order terms dominates. Thus, approximating $f(X, Y)$ with $\frac{\mathbb{E}[X]}{\mathbb{E}[X]+\mathbb{E}[Y]+an}$ results in an approximation error of
    \begin{equation*}
        \left|\frac{X}{X+Y+an} - \frac{\mathbb{E}[X]}{\mathbb{E}[X]+\mathbb{E}[Y]+an}\right| = \mathcal{O}_p\left(\frac{1}{\sqrt{n}}\right).  
    \end{equation*}
    
    We have now shown that
    \begin{equation*}
        \frac{X}{X+Y+an} \xrightarrow{~p~}  \frac{\mathbb{E}[X]}{\mathbb{E}[X]+\mathbb{E}[Y]+an},    
    \end{equation*}
    when $n \rightarrow \infty$. Since for all $n$, the expression $\frac{X}{X+Y+an}$ is bounded by $0\leq\frac{X}{X+Y+an}\leq 1$, it follows from the Lebesgue Dominated Convergence Theorem that
    \begin{equation*}
        \mathbb{E}\left[\frac{X}{X+Y+an}\right] \rightarrow \mathbb{E}\left[\frac{\mathbb{E}[X]}{\mathbb{E}[X]+\mathbb{E}[Y]+an}\right] = \frac{\mathbb{E}[X]}{\mathbb{E}[X]+\mathbb{E}[Y]+an},    
    \end{equation*}
    when $n \rightarrow \infty$ and that the approximation error decays at the rate of $\mathcal{O}\left(\frac{1}{\sqrt{n}}\right)$.
\end{proof}

\section{Additional Algorithms}\label{app:algos}

Algorithm~\ref{alg:hdi} is guaranteed to find the $(1-\alpha)$ HDI, where $\alpha$ is a user set parameter ($\alpha=0.05$ corresponds with a 95\% confidence interval). It uses a two-pointer approach to narrow the interval to a minimum length such that the probability mass within that interval is at least $(1-\alpha)\cdot 100\%$ of the whole distribution. The while-loop within the algorithm takes at most $\mathcal{O}(n)$ iterations, but it relies on the associative array being sorted by key. The worst-case time complexity of the sorting operation is $\mathcal{O}(n\log n)$, which becomes the time complexity of the whole algorithm since all other operations can be done in constant time.

\begin{algorithm}[ht]
\caption{HDI}\label{alg:hdi}
\begin{algorithmic}[1]
\Require $X_{metric}, \alpha$ \quad\quad\quad(Distribution for a given metric and a confidence level $\alpha$)
\Ensure $(lower, upper)$ \quad (Bounds for the $1-\alpha$ HDI)
\State $X_{sorted} \gets sort(X_{metric})$
\State $l \gets 0$
\State $u \gets |X_{sorted}|-1$
\State $tail\_coverage \gets 0$
\State $bounds\_not\_found \gets True$
\While{$bounds\_not\_found$}
    \State $p_l \gets X_{sorted}[l][1]$
    \State $p_u \gets X_{sorted}[u][1]$
    \If{$p_l < p_u$}
        \If{$tail\_coverage + p_l < \alpha$}
            \State $tail\_coverage \gets tail\_coverage + p_l$
            \State $l \gets l + 1$
        \Else
            \State $bounds\_not\_found \gets False$
        \EndIf
    \Else
        \If{$tail\_coverage + p_u < \alpha$}
            \State $tail\_coverage \gets tail\_coverage + p_u$
            \State $u \gets u - 1$
        \Else
            \State $bounds\_not\_found \gets False$
        \EndIf
    \EndIf
\EndWhile
\State $lower \gets X_{sorted}[l][0]$
\State $upper \gets X_{sorted}[u][0]$
\end{algorithmic}
\end{algorithm}

\section{Additional experiments}\label{app:experiments}

\subsection{Simulated covariate shift}

As discussed in Section~\ref{sec:discussion}, confidence-based estimators are not guaranteed to work under concept shift, which is present in current binary classification benchmarks such as TableShift~\citep{gardner:2024}. Thus, we sought to demonstrate the effectiveness of CBPE in a situation where we could control the precise nature of the shift in the data distribution. For this, we created synthetic data where the concepts remain stable and only covariate shift is present.

To further demonstrate that the effectiveness of CBPE is not dependent on the number of features, we created several synthetic datasets with different numbers of features. In all cases, we first chose $n$ as the number of features. We then sampled points from the $n$-dimensional (Euclidean) space and assigned labels to each sampled point with the following logic. All points that fell on the $n$-dimensional hypersphere with radius 3, were assigned with the positive label. For all other points, the probability of being assigned the positive label depended on their distance $d$ from the hypersphere by the scaled sigmoid function $f(d)=e^{-\lambda d^2}$, where $\lambda$ is a hyperparameter controlling the rate of change in the probabilities. We used $\lambda=\ln{\sqrt{2}}$ in all experiments for a suitable change in the label probability. This data generation process was designed to be both simple and yet able to create a challenging non-linear decision boundary that would make the learning task non-trivial for ML models.

The points were sampled from two pools, where one pool had points either very close or relatively far from the hypersphere. These points were assigned with labels 0 or 1 with high probability and could be considered easy to predict. The other pool consisted of points lying in the in-between space where label entropy was maximized and each point was roughly equally likely to be assigned with either of the labels. These points could be considered hard to predict. Our sampling procedure also created a slight bias in the label distribution to ensure that our results would not be restricted only to cases where the dataset was balanced. 

\begin{figure}[ht!]
\centering
\includegraphics[width=1.0\columnwidth]{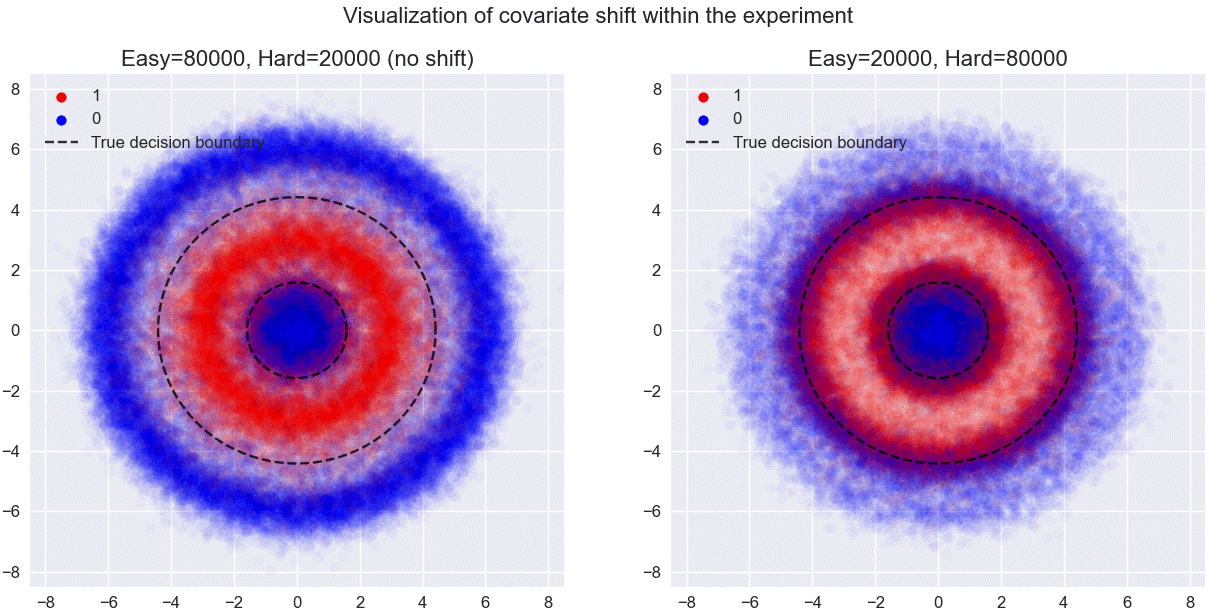}
\caption{Vizualization of our synthetic dataset for the 2-dimensional case.}
\label{fig:dataset}
\end{figure}

We simulated the covariate shift by sampling two sets of 100,000 points for each $n$ used in our trials (we used $n = 2, 10, 100, 1000$). The first set had 80\% of easy-to-predict points and 20\% of hard-to-predict points. We trained an XGBoost model using 60\% of the first set. We then used another 20\% to train a calibration mapping using Isotonic Regression. The remaining 20\% was used as an ID-test set in evaluating the CBPE performance. For the second set, the easy- and hard-to-predict proportions were swapped to simulate covariate shift in the dataset. Since the conditional label probabilities remained the same, no concept shift was introduced. The whole second set was used as an OoD test set. The resulting datasets for the two-dimensional feature space are demonstrated in Figure~\ref{fig:dataset}. The dashed line presents the decision boundary for the Bayes optimal classifier (where $p(y=1\mid\boldsymbol{x})=0.5$). For higher dimensional feature spaces, the structure of the datasets is analogical.

For both test sets, we ran 1,000 trials with a monitoring window size of 1,000 as well. In each trial, we drew a random sample from the test set, estimated the performance using the four metrics of our interest, and compared those against the actual values of the metrics over the whole test set. The results are shown in Table~\ref{tab:covariate}, where for both test sets we report class imbalance (the prevalence of the positive class) as "Bias", calibration error of each model as "ACE", the actual values for the four metrics of interest, and the mean absolute estimation errors for the CBPE method as "MAE". The error bounds for the estimation errors are given in the form of two standard deviations. All values are percentages except error bounds, which are given as percentage units.  

\begin{table*}[!ht]
    \footnotesize
    \centering
    \caption{Mean Absolute Estimation Errors for Different Metrics with Simulated data (\%)}
    \setlength{\tabcolsep}{4.2pt}
    \begin{tabular}{l|c|c|c|c|c|c|c|c|c|c|c}
    Fea- & \multicolumn{3}{c}{} & \multicolumn{2}{|c}{Accuracy} & \multicolumn{2}{|c}{Precision} & \multicolumn{2}{|c}{Recall} & \multicolumn{2}{|c}{F$_1$} \\
        tures & Split & Bias & ACE & True & MAE & True & MAE & True & MAE & True & MAE \\ \hline
        \multirow{2}{*}{2} & ID & 32.7 & 0.71 & 87.5 & -0.16$\pm$12 & 84.9 & 1.35$\pm$12 & 75.2 & -2.44$\pm$12 & 79.8 & -0.83$\pm$12 \\ 
        & OoD & 43.8 & 2.40 & 71.7 & 2.08$\pm$11 & 69.0 & 5.00$\pm$11 & 64.1 & -3.65$\pm$11 & 66.5 & 0.07$\pm$11 \\ \hline
        \multirow{2}{*}{10} & ID & 40.1 & 0.84 & 83.2 & 0.24$\pm$5.7 & 80.5 & 0.90$\pm$5.7 & 76.9 & -0.66$\pm$5.7 & 78.7 & 0.07$\pm$5.7 \\ 
        & OoD & 48.0 & 1.03 & 76.1 & 0.12$\pm$3.9 & 75.7 & 0.95$\pm$3.9 & 73.9 & -1.59$\pm$3.9 & 74.8 & -0.38$\pm$3.9 \\ \hline
        \multirow{2}{*}{100} & ID & 40.2 & 1.04 & 83.4 & 0.03$\pm$4.3 & 80.2 & 0.13$\pm$4.3 & 77.9 & 0.23$\pm$4.3 & 79.0 & 0.17$\pm$4.3 \\ 
        & OoD & 47.7 & 1.31 & 75.7 & 1.09$\pm$2.2 & 75.3 & 0.56$\pm$2.2 & 73.1 & 1.85$\pm$2.2 & 74.2 & 1.21$\pm$2.2 \\ \hline
        \multirow{2}{*}{1000} & ID & 40.1 & 0.74 & 83.3 & -0.30$\pm$4.9 & 80.2 & -0.14$\pm$4.9 & 77.5 & -0.75$\pm$4.9 & 78.9 & -0.47$\pm$4. \\ 
        & OoD & 47.8 & 1.77 & 74.6 & 1.74$\pm$2.5 & 73.9 & 1.78$\pm$2.5 & 72.4 & 1.39$\pm$2.5 & 73.2 & 1.58$\pm$2.5 \\ \hline
    \end{tabular}
    \label{tab:covariate}
\end{table*}

The results show that on average CBPE results in rather small estimation errors both for the ID and OoD data on all metrics. Particularly, the mean estimation errors are much smaller than the change in the values of the metrics induced by the covariate shift, which makes detecting persistent changes in the underlying data distribution possible. These results hold for varying feature space dimensions and label biases. The deviation in the estimates is relatively high in some cases, but this is somewhat inherent in all estimation methods. The only way to combat this is to increase the monitoring window size. 

\end{appendices}

\bibliography{references}

\end{document}